\documentclass[sigconf]{acmart}

\renewcommand\footnotetextcopyrightpermission[1]{}
\usepackage{multirow}
\usepackage{subcaption}
\usepackage[stable]{footmisc}

\AtBeginDocument{%
  \providecommand\BibTeX{{%
    \normalfont B\kern-0.5em{\scshape i\kern-0.25em b}\kern-0.8em\TeX}}}

\acmConference[DiG]{}{2022}{ArXiv}

\begin{document}

\title{Reading and Writing: Discriminative and Generative Modeling for Self-Supervised Text Recognition}

\author{Mingkun Yang}
\authornote{Both authors contributed equally to this research.}
\affiliation{
    \institution{Huazhong University of Science and Technology}
    \country{}}
\email{yangmingkun@hust.edu.cn}

\author{Minghui Liao}
\authornotemark[1]
\affiliation{
    \institution{Huawei Inc.}
    \country{}}
\email{mhliao@foxmail.com}

\author{Pu Lu}
\affiliation{
    \institution{Huawei Inc.}
    \country{}}
\email{lupu1@huawei.com}

\author{Jing Wang}
\affiliation{
    \institution{Huawei Inc.}
    \country{}}
\email{wangjing105@huawei.com}

\author{Shenggao Zhu}
\affiliation{
    \institution{Huawei Inc.}
    \country{}}
\email{zhushenggao@huawei.com}

\author{Hualin Luo}
\affiliation{
    \institution{Huawei Inc.}
    \country{}}
\email{luohualin@huawei.com}

\author{Qi Tian}
\affiliation{
    \institution{Huawei Inc.}
    \country{}}
\email{tian.qi1@huawei.com}

\author{Xiang Bai}
\affiliation{
    \institution{Huazhong University of Science and Technology}
    \country{}}
\email{xbai@hust.edu.cn}

\renewcommand{\shortauthors}{Mingkun Yang and Minghui Liao, et al.}

\begin{abstract}
Existing text recognition methods usually need large-scale training data. Most of them rely on synthetic training data due to the lack of annotated real images. However, there is a domain gap between the synthetic data and real data, which limits the performance of the text recognition models. Recent self-supervised text recognition methods attempted to utilize unlabeled real images by introducing contrastive learning, which mainly learns the discrimination of the text images. Inspired by the observation that humans learn to recognize the texts through both reading and writing, we propose to learn discrimination and generation by integrating contrastive learning and masked image modeling in our self-supervised method. The contrastive learning branch is adopted to learn the discrimination of text images, which imitates the reading behavior of humans. Meanwhile, masked image modeling is firstly introduced for text recognition to learn the context generation of the text images, which is similar to the writing behavior. 
The experimental results show that our method outperforms previous self-supervised text recognition methods by 10.2\%-20.2\% on irregular scene text recognition datasets. Moreover, our proposed text recognizer exceeds previous state-of-the-art text recognition methods by averagely 5.3\% on 11 benchmarks, with similar model size. We also demonstrate that our pre-trained model can be easily applied to other text-related tasks with obvious performance gain. The code is available at \url{https://github.com/ayumiymk/DiG}.
\end{abstract}

\maketitle

\section{Introduction}

Reading text from images~\cite{ijcv/JaderbergSVZ16,LiaoPHHB20} is a long-standing and valuable topic that bridges vision and language, which mainly consists of text detection~\cite{tip/LiaoSB18,aaai/LiaoWYCB20} and text recognition~\cite{crnn,aster}. Since most existing text recognition methods are data-hungry while the annotated real images are expensive, they usually rely on large-scale synthetic training data. However, the domain gap between the synthetic data and real data limits the performance of the text recognition models. Thus, it is meaningful to explore the usage of unlabeled real images, which can further inspire the potential of the text recognition models.

\begin{figure}[t]
\centering
\captionsetup[subfigure]{justification=centering}
\begin{subfigure}[b]{0.48\textwidth}
         \centering
         \includegraphics[width=\textwidth]{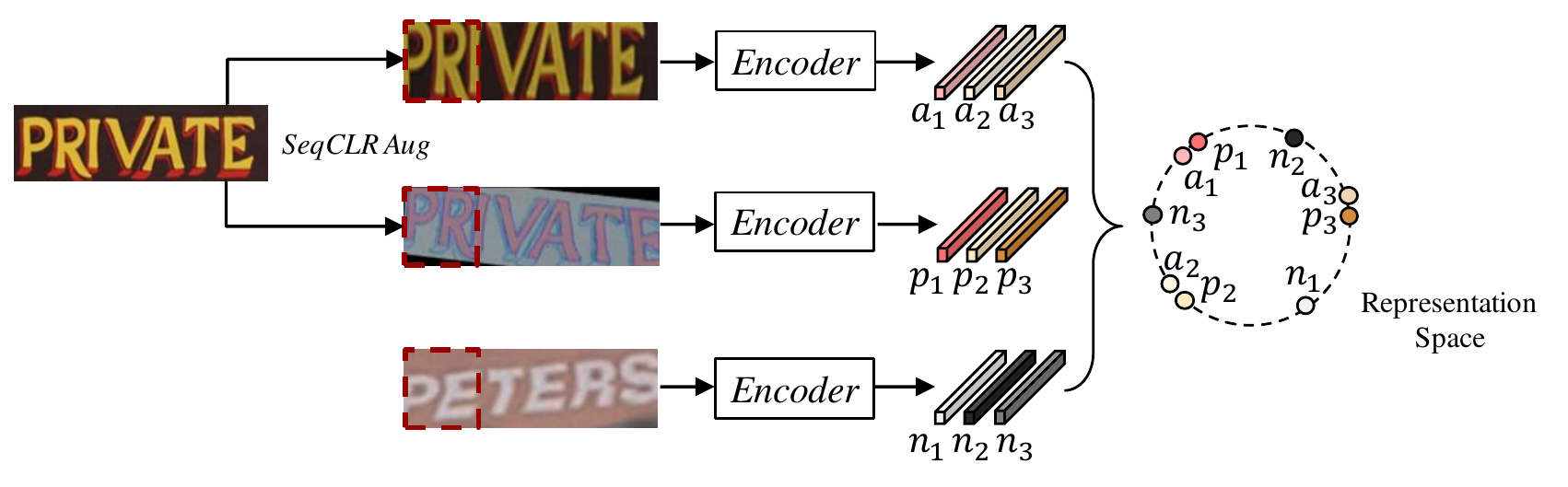}
         \caption{}
         \label{subfig:introduction1}
\end{subfigure}
\begin{subfigure}[b]{0.48\textwidth}
         \centering
         \includegraphics[width=\textwidth]{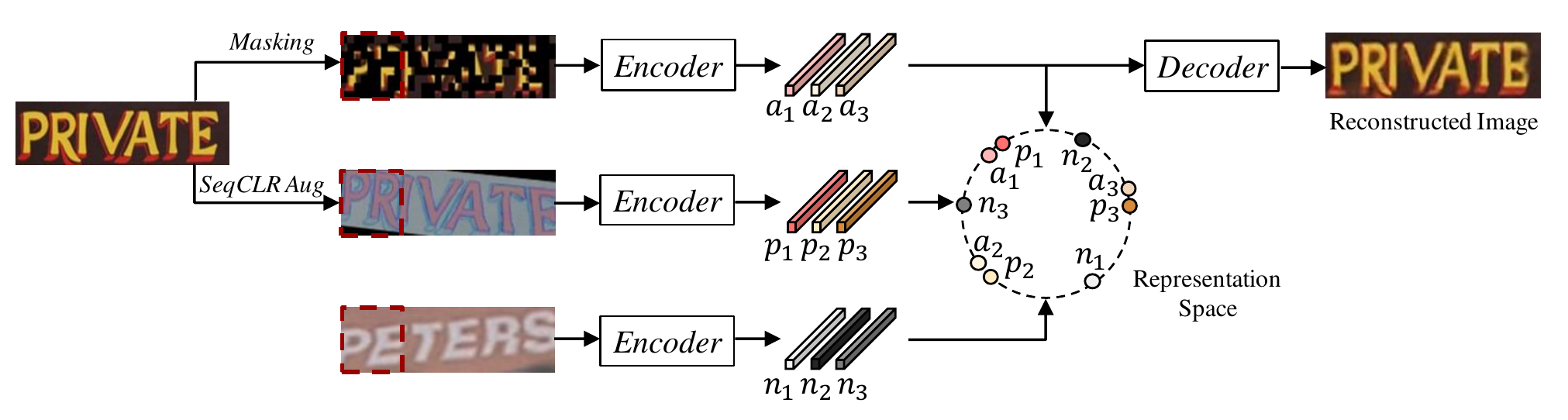}
         \caption{}
         \label{subfig:introduction2}
\end{subfigure}
\caption{Illustrations of (a) self-supervised text recognition methods based on contrastive learning and (b) our proposed DiG that integrates contrastive learning and masked image modeling.}
\vspace{-4mm}
\label{fig:introduction}
\end{figure}

It is natural to adopt self-supervised learning for text recognition to sufficiently make use of real images. Previous works attempted to utilize unlabeled real images by introducing contrastive learning. SeqCLR~\cite{SeqCLR} proposed a sequence-to-sequence contrastive learning framework for text recognition. PerSec~\cite{liu2022perceiving} introduced a hierarchical contrastive learning method for text recognition. They are both based on contrastive learning and mainly focus on learning the discrimination of text images, as shown in Fig.~\ref{subfig:introduction1}.
We observe that our humans learn to recognize text images by both reading and writing. Reading means observing the text of different appearances or from different viewpoints, which helps us learn the discrimination. Writing is a generative manner that goes deeper into recognizing text images. The combination of reading and writing helps humans to recognize the text better. Inspired by this observation, we propose a Discriminative and Generative self-supervised method (DiG) for text recognition.

As shown in Fig.~\ref{subfig:introduction2}, our proposed DiG integrates contrastive learning and masked image modeling into a unified model, to sufficiently enjoy the superiority of the discriminative model and the generative model. Specifically, two views of the input image named masked view and augmented view, are fed into the encoder, to perform contrastive learning. Meanwhile, the masked view is additionally used for masked image modeling. In this way, DiG can learn both the discrimination and the generation of text images, which produces more robust feature representations for text images.

With the assistance of our proposed DiG, we can pre-train foundation models for text recognition, as well as some other text-related tasks, such as text segmentation and text image super-resolution.
The contributions of this paper are three-fold:
 
(1) We propose a discriminative and generative self-supervised model, which is the first to introduce generative self-supervised learning for text recognition.

(2) Our proposed DiG integrates contrastive learning and masked image modeling into a unified model, taking advantage of learning both discrimination and generation. The experiments of feature representation evaluation show that DiG outperforms the generative model and the discriminative model by averagely 11.8\% and 4.1\%, respectively.

(3) Our DiG can significantly improve the accuracy of various text-related tasks, including text recognition, text segmentation, and text image super-resolution. Concretely, our text recognition models not only surpass previous self-supervised text recognition methods by 10.2\%-20.2\% on irregular scene text recognition datasets but also outperform previous state-of-the-art text recognition methods by averagely 5.3\% on 11 benchmarks, with similar model size.

\section{Related Work}

\subsection{Self-Supervised Learning}
Recent self-supervised learning methods for vision can be roughly divided into two categories: contrastive learning and masked image modeling, which are the representative techniques of discriminative methods and generative methods, respectively. 

\noindent \textbf{Contrastive Learning} Contrastive learning methods~\cite{HadsellCL06,DosovitskiyFSRB16} learned the visual representation with positive pairs and negative pairs. MoCo~\cite{moco} built a dynamic dictionary with a queue and a moving-averaged encoder, which treated contrastive learning as dictionary look-up. SimCLR~\cite{SimCLR} simplified contrastive learning algorithms by removing specialized architectures or a memory bank. MoCo v2~\cite{mocov2} extended MoCo by adopting two effective designs in SimCLR, which further demonstrates the superiority of MoCo. SimCLRv2~\cite{SimCLRv2} presented a new three-step pipeline, including unsupervised or self-supervised pretraining, supervised fine-tuning, and distillation using unlabeled data. Different from previous methods that mostly focused on CNNs, MoCo v3~\cite{mocov3} investigated applying contrastive learning for vision transformers by alleviating the instability issues.

\noindent \textbf{Masked Image Modeling} Masked image modeling~\cite{DoerschGE15,ChenRC0JLS20} has developed rapidly recently, in parallel with the development of masked language modeling~\cite{bert} in natural language processing. MAE~\cite{mae} and SimMIM~\cite{simMIM} proposed masked image modeling algorithms for vision transformers. They masked random patches of the input image and reconstructed the missing pixels. MaskFeat~\cite{maskfeat} investigated the reconstructing target and replaced the pixels with HoG~\cite{hog} features for reconstruction. Compared to the contrastive learning methods, masked image modeling provided another view for self-supervised learning and was simple yet effective.

\subsection{Text Recognition}
Recent text recognition methods include sequence-to-sequence~\cite{crnn,aster,YangGLHBBYB19} and segmentation-based~\cite{LiaoZWXLLYB19} approaches, where the former is more flexible and convenient profiting from their end-to-end manner.
Existing text recognition methods based on sequence-to-sequence models can be classified into three types according to the decoders: CTC decoder, attention decoder, and Transformer decoder. CRNN~\cite{crnn} was a representative CTC-based text recognition method. It consisted of a CNN feature extractor, an RNN sequence encoder, and a CTC decoder. Most of the CTC-based text recognition methods~\cite{conf/aaai/HeH0LT16,su2017accurate} adopted similar pipelines. Attention decoders were also popular in previous text recognition methods. For example, Shi et al.~\cite{aster} designed a text recognition model with an attention decoder and a spatial transform network for irregular text recognition. Spatial attention decoders were further adopted in Yang et al.~\cite{yang2017learning} and Wojna et al.~\cite{wojna2017attention}. Recently, owing to the rapid rising of Transforms, some text recognition methods~\cite{master,satrn} applied Transformer decoders for text recognition models.

\subsection{Self-Supervised Text Recognition}
Sequence-to-sequence models have dominated the text recognition area in recent years since they do not need character-level annotations. Thus, existing self-supervised text recognition methods are based on sequence-to-sequence models. SeqCLR~\cite{SeqCLR} proposed a sequence-to-sequence contrastive learning framework for text recognition. It applied contrastive learning to the individual elements of the sequence while keeping the information of their order. The proposed framework was effective with both CTC decoders and attention decoders.
PerSec~\cite{liu2022perceiving} also introduced contrastive learning for text recognition. It presented hierarchical contrastive learning that drove each element of features at high and low levels. PerSec was pre-trained with 100M unlabeled images and fine-tuned with synthetic data to demonstrate its effectiveness.

Compared to these methods that only applied contrastive learning for text recognition, we integrate contrastive learning and masked image modeling into a unified model, sufficiently enjoying the advantages of both discriminative learning and generative learning.

\section{Methodology}

\begin{figure*}[ht]
\centering
\includegraphics[width=0.98\linewidth]{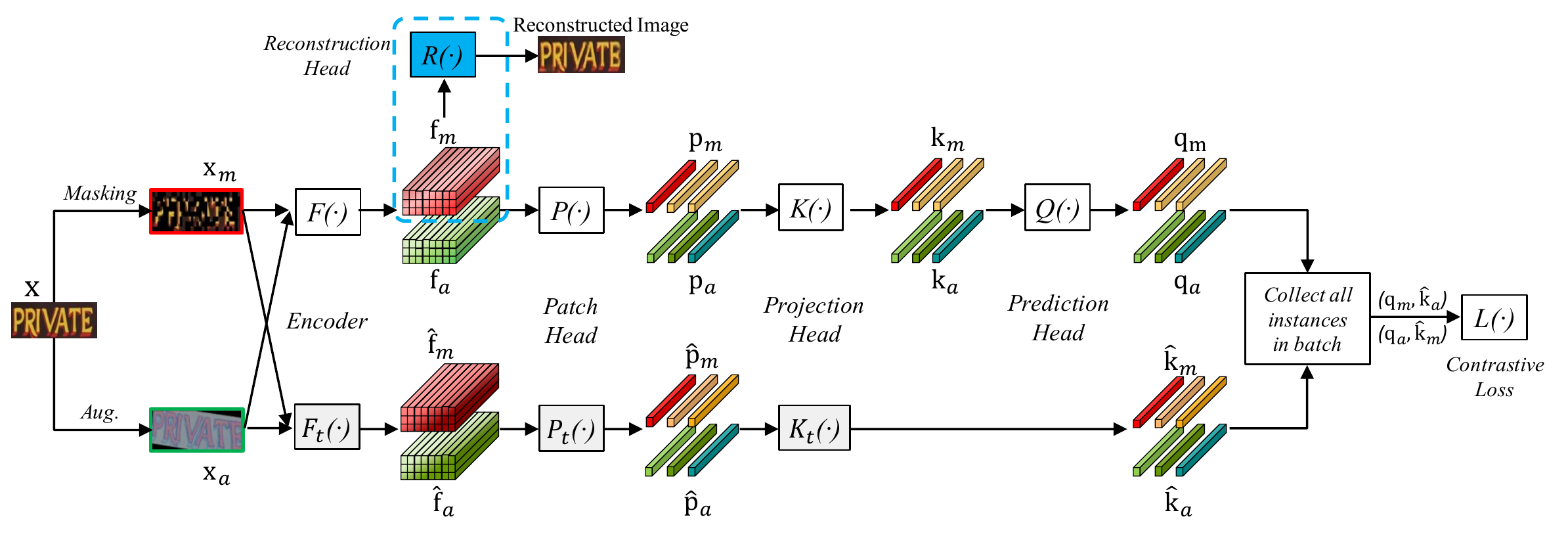}
\caption{Architecture of DiG. The red ingredients are features of the masked view from different stages while the green items are relevant to the augmented view. Text boxes without any background mean modules of the online branch while the ones with grey backgrounds represent modules of the momentum branch.}
\vspace{-2mm}
\label{fig:methodology}
\end{figure*}

We follow the common pipeline of self-supervised learning which includes self-supervised pre-training and task-specific fine-tuning.
In the task-specific fine-tuning period, the model consists of an encoder trained from the self-supervised pre-training period and a task-specific decoder.

We first introduce the architecture of our proposed method DiG in Sec.~\ref{subsec:architecture}. Then, the contrastive learning module and masked image modeling module are depicted in Sec.~\ref{subsec:cl} and Sec.~\ref{subsec:mim} respectively. After that, the details of the optimization are described in Sec.~\ref{subsec:optimization}. Finally, we present three down-stream tasks including text recognition (Sec.~\ref{subsec:tr}), text image super-resolution (Sec.~\ref{subsec:tisr}), and text segmentation (Sec.~\ref{subsec:ts}).

\subsection{Architecture}\label{subsec:architecture}
The architecture of our proposed DiG is shown in Fig.~\ref{fig:methodology}. The input image is firstly resized to $H \times W$, where $H=32$ and $W=128$ are the height and width of the input image. Then, two views of the input image, including the masked view $X_m$ and the augmented view $X_a$, are fed into a ViT~\cite{vit} encoder $F(\cdot)$, producing the masked feature $f_m$ and the augmented feature $f_a$, respectively. In the masked image modeling branch, a reconstruction head is applied to $f_m$ to reconstruct the target of masked image modeling. In the contrastive learning branch, $f_m$ and $f_a$ are processed by the patch head $P(\cdot)$, projection head $K(\cdot)$, and prediction head $Q(\cdot)$ to form the corresponding queries $q_m$ and $q_a$. Besides, a momentum branch that consists of $F_t(\cdot)$, $P_t(\cdot)$, and $K_t(\cdot)$ is utilized to produce keys $\hat{k}_m$ and $\hat{k}_a$. Finally, queries and keys (including the negative samples from other images, which are not shown in Fig.~\ref{fig:methodology}) are collected to calculate the contrastive loss.

\subsection{Contrastive Learning}\label{subsec:cl}
Our contrastive learning algorithm is inherited from MoCo v3~\cite{mocov3} with some minor adaptations. The details of our contrastive branch are as follows.

\noindent \textbf{Data augmentation} For effective contrastive learning, the data augmentation plays an essential role in the whole framework. Following~\cite{liu2022perceiving}, besides the augmentation operations used in SeqCLR, more image augmentations including color jitter and gray are employed to enhance the representation quality of the pre-trained model.

\noindent \textbf{Encoder} We use a ViT~\cite{vit} as the feature encoder. Specifically, the input image is firstly split into non-overlapping patches with size $4 \times 4$, and then embedded via a linear mapping layer with added positional embeddings. The resulting patches are processed using a stack of Transformer blocks.

\noindent \textbf{Patch Head} Most existing contrastive learning methods are designed for general object images. Usually, the encoded feature maps are projected into a single vector, which is then used as the atomic element in a contrastive loss. However, the text image is a sequence-like object, whose feature maps represent a series of characters. Following SeqCLR~\cite{SeqCLR}, the encoded feature maps are horizontally split into four patches empirically and each patch is considered as the atomic input element for contrastive learning.

\noindent \textbf{Projection Head} The projection head $K(\cdot)$ is a 3-layer MLP, where each layer has a fully-connected (fc) layer and a layer normalization. GELU is also applied to each layer except the output fc. The hidden fc is 4096-d while the output fc is 256-d.

\noindent \textbf{Prediction Head} The prediction head $Q(\cdot)$ has a similar architecture with $K(\cdot)$ but only consists of 2 MLP layers.

\noindent \textbf{Momentum Branch} To facilitate contrastive learning via building a large and consistent dictionary on-the-fly, a momentum branch is adopted. This branch is composed of $F_t(\cdot)$, $P_t(\cdot)$, and $K_t(\cdot)$, which have the same architecture with $F(\cdot)$, $P(\cdot)$, and $K(\cdot)$ and are parameterized by an exponentially moving average (EMA) of their parameters respectively.

\subsection{Masked Image Modeling}\label{subsec:mim}
Our masked image modeling module follows SimMIM~\cite{simMIM}, which consists of the following four major components.

\noindent \textbf{Masking Strategy} We adopt a patch-aligned random masking strategy. The patch size is set to $4 \times 4$. The mask ratio is set to 0.6 empirically. Each masked patch is replaced with a learnable mask token vector.

\noindent \textbf{Encoder} We share the ViT~\cite{vit} encoder with the contrastive learning branch.

\noindent \textbf{Prediction Head} Following SimMIM, we use a linear layer as the prediction head, which is extremely lightweight.

\noindent \textbf{Reconstruction Target} We adopt the RGB values of the raw pixels as the reconstruction target. 

\subsection{Optimization}\label{subsec:optimization}
The loss function $L$ of our proposed DiG is formulated as:
\begin{equation}
    L = L_c + \alpha \times L_m, 
\end{equation}
where $L_c$ and $L_m$ indicate the loss function of contrastive learning and the loss function of masked image modeling, respectively. $\alpha$ is a scaling factor, which is set to 0.1 empirically.

We apply an InfoNCR~\cite{infonce} loss for $L_c$. Assuming the two queries from the masked view and the augmented view are $q_m$ and $q_a$ and their corresponding keys from the momentum branch are $\hat{k}_m$ and $\hat{k}_a$, the contrastive loss $L_c$ can be formulated as:
\begin{equation}
\begin{aligned}
    L_c =  &-log\frac{exp(q_m \cdot \hat{k}_a / \tau)}{exp(q_m \cdot \hat{k}_a / \tau) + \sum_{\hat{k}_a^-}{exp(q_m \cdot \hat{k}_a^- / \tau)}} \\
           &- log\frac{exp(q_a \cdot \hat{k}_m / \tau)}{exp(q_a \cdot \hat{k}_m / \tau) + \sum_{\hat{k}_m^-}{exp(q_a \cdot \hat{k}_m^- / \tau)}},
\end{aligned}
\end{equation}
Here $\hat{k}_m^-$ indicates the negative samples obtained from other input images in the same batch. $\tau$ denotes the temperature, which is set to $0.2$.

An $L2$ loss function is employed for $L_m$:
\begin{equation}
    L_m = \frac{1}{N} \sum_{i \in N}{(x_i - y_i)^2}, 
\end{equation}
where $x_i, y_i \in R^{3}$ are the predicted values and the RGB targets of masked pixel $i$; $N$ is the number of masked pixels.

\subsection{Text Recognition}\label{subsec:tr}
\begin{figure}[ht]
\centering
\includegraphics[width=0.98\linewidth]{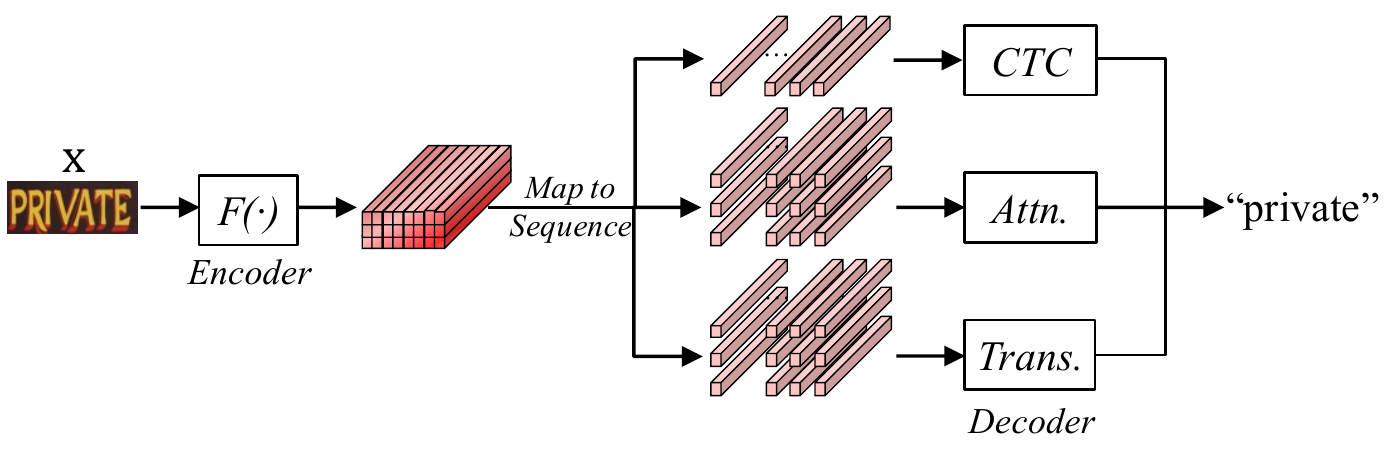}
\caption{Pipeline of our text recognition method.}
\vspace{-2mm}
\label{fig:text-recognition}
\end{figure}

Our text recognition model consists of a ViT encoder and a text recognition decoder, where the encoder is inherited from DiG and the decoder converts the 2D features to a sequence of characters. The text recognition decoder can be CTC decoder~\cite{crnn}, attention decoder~\cite{aster}, or Transformer decoder~\cite{transformer,satrn}.
The Transformer decoder is composed of a stack of 6 identical layers, which is the same as the recent ViT-based text recognition method SATRN~\cite{satrn}. Since CTC decoder is conducted on the 1D feature sequence, the encoded feature maps are vertically average-pooled before being fed into the subsequent decoder. The workflow of text recognition is illustrated in Fig.~\ref{fig:text-recognition}.

\subsection{Text Segmentation}\label{subsec:ts}
Our text segmentation model is composed of a ViT encoder inherited from DiG and a lightweight text segmentation decoder. The text segmentation decoder consists of 3 multi-head attention layers and a linear prediction head, where the number of heads is set to 2 and the embedding dimension is 384. An L1 loss is applied to the text segmentation model.

\subsection{Text Image Super-Resolution}\label{subsec:tisr}
Our text image super-resolution model contains a ViT encoder from DiG and a lightweight super-resolution decoder. The super-resolution decoder is similar to the text segmentation decoder. We adopt an L2 loss for this task.

\section{Experiments}
\subsection{Datasets}\label{sec:datasets}
\noindent \textbf{Unlabeled Real Data (URD)} To alleviate the domain gap between realistic test data and synthetic training data, we collect about 15.77M unlabeled text images (termed CC-OCR) from Conceptual Captions Dataset (CC)~\footnote{\url{https://github.com/google-research-datasets/conceptual-captions}}, which are used for self-supervised learning. Specifically, CC is a large-scale image-text dataset that is usually used for VL pre-training. In TAP~\cite{yang2021tap}, the authors released the OCR results of CC obtained by Microsoft Azure OCR system. Therefore, we can get text image patches via the given OCR results and CC images.

\noindent \textbf{Synthetic Text Data (STD)} Our synthetic text data is a combination of SynthText~\cite{synthtext} that contains 8M text boxes from 800k images and Synth90k~\cite{synth90k} that contains 9M images for text recognition. Thus, the synthetic text data include 17M images for text recognition in total.

\noindent \textbf{Annotated Real Data (ARD)} We collect about 2.78M annotated text images from TextOCR~\cite{textocr} (0.71M) and Open Images Dataset v5~\footnote{\url{https://storage.openvinotoolkit.org/repositories/openvino_training_extensions/datasets/open_images_v5_text}} (2.07M). 

\noindent \textbf{Scene Text Recognition Benchmarks} We evaluate our text recognition models on the following scene text recognition benchmarks:
(1) IIIT5k-Words (IIIT)~\cite{DBLP:conf/cvpr/MishraAJ12}; (2) Street View Text (SVT)~\cite{wang2011end}; (3) ICDAR 2013 (IC13)~\cite{karatzas2013icdar}; (4) ICDAR 2015 Incidental Text (IC15)~\cite{karatzas2015icdar}; (5) SVTP-Perspective (SVTP)~\cite{quy2013recognizing}; (6) CUTE80 (CUTE)~\cite{risnumawan2014robust}; (7) COCO-Text-Validation (COCO)~\cite{cocotext}; (8) CTW dataset~\cite{ctw1500}; (9) Total-Text dataset (TT)\cite{CK2017}; (10) Occlusion Scene Text (OST)~\cite{ost}, including weakly occluded scene text(WOST) and heavily occluded scene text (HOST). The descriptions of the benchmarks are shown in Tab.~\ref{tab:benchmarks}.

\noindent \textbf{Handwritten Text Recognition Benchmarks} We verify our text recognition models on two handwritten text recognition benchmarks: CVL~\cite{CVL} and IAM~\cite{IAM}. Following SeqCLR~\cite{SeqCLR} and PerSec~\cite{liu2022perceiving}, we collect 146805 images for training from CVL and IAM. 12012 images and 13752 images are assigned to the test set of CVL and IAM, respectively. 

\begin{table}[ht]
\centering
\caption{Descriptions of Scene Text Recognition Benchmarks.}
\label{tab:benchmarks}
\begin{tabular}{|c|c|c|}
\hline
Benchmarks & \#test images & Description \\ \hline
IIIT           &   3000            &  regular scene text           \\ \hline
SVT           &    647           &   regular scene text          \\ \hline
IC13           &   1015            &   regular scene text          \\ \hline
COCO           &  9896             &   incidental scene text          \\ \hline
IC15           &   1811            &   incidental scene text          \\ \hline
SVTP           &   645            &    perspective scene text         \\ \hline
CUTE           &   288            &    irregular scene text         \\ \hline
CTW           &   1572            &    irregular scene text         \\ \hline
TT           &    2201           &     irregular scene text        \\ \hline
WOST           &   2416            &    weakly occluded scene text         \\ \hline
HOST           &   2416            &    heavily occluded scene text         \\ \hline

\end{tabular}
\end{table}

\noindent \textbf{Text Segmentation Benchmark} TextSeg~\cite{textseg} is a fine-annotated text dataset for text segmentation. It contains 4024 images (training set: 2646, validation set: 340, testing set: 1038).

\noindent \textbf{Text Image Super-Resolution Benchmark} TextZoom dataset~\cite{textzoom} is a paired scene text super-resolution dataset. The training set contains 17367 pairs of images. The test dataset is split into three parts (easy, medium, and hard) according to the difficulty, which is composed of 1619, 1411, and 1343 pairs of images, respectively.

\begin{table*}[ht]
\caption{Feature representation evaluation of DiG on scene text recognition benchmarks. ``Gen-$\ast$": masked image modeling for self-supervised pre-training; ``Dis-$\ast$": contrastive learning for self-supervised pre-training; ``DiG-$\ast$": integration of masked image modeling and contrastive learning. ``Avg." means per-image accuracy on all the scene text recognition benchmarks.}
\label{tab:probe-eval}
\resizebox{0.99\textwidth}{!}{
\begin{tabular}{lcccccccccccc}
\hline
\multicolumn{1}{c}{\multirow{2}{*}{Method}}     & \multicolumn{3}{c}{Regular} & \multicolumn{6}{c}{Irregular}            & \multicolumn{2}{c}{Occluded} & \multicolumn{1}{c}{\multirow{2}{*}{Avg.}}  \\
\cmidrule(lr){2-4} \cmidrule(lr){5-10} \cmidrule(lr){11-12}
\multicolumn{1}{c}{}                        & IIIT    & SVT     & IC13    & IC15 & SVTP & CUTE & COCO & CTW  & TT   & HOST & WOST                      &                       \\ \hline
\multicolumn{1}{l|}{Gen-ViT-Small}          & 86.6    & 82.1    & 88.7    & 72.9 & 74.4 & 72.2 & 48.5 & 64.1 & 63.3 & 33.8 & \multicolumn{1}{c|}{56.5} & 59.3                  \\
\multicolumn{1}{l|}{Dis-ViT-Small}          & 92.6    & 90.4    & 93.4    & 81.2 & 81.7 & 84.0 & 60.0 & 72.8 & 73.1 & 33.3 & \multicolumn{1}{c|}{56.1} & 67.0                  \\
\multicolumn{1}{l|}{DiG-ViT-Small}          & \textbf{94.2}    & \textbf{93.0}    & \textbf{95.3}    & \textbf{84.3} & \textbf{86.1} & \textbf{87.5} & \textbf{63.4} & \textbf{77.9} & \textbf{75.8} & \textbf{41.7} & \multicolumn{1}{c|}{\textbf{64.0}} & \textbf{71.1}                  \\ \hline
\end{tabular}
}
\end{table*}

\begin{table*}[ht]
\caption{Fine-tuning evaluation on scene text recognition benchmarks. ``Label Fraction" means training with different percentages of the annotated real data. ``Scratch-$\ast$": without self-supervised pre-training.}
\label{tab:ablation-cl-mim}
\resizebox{0.99\textwidth}{!}{
\begin{tabular}{cccccccccccccc}
\hline
\multicolumn{1}{l}{\multirow{2}{*}{Label Fraction}} & \multicolumn{1}{c}{\multirow{2}{*}{Method}}     & \multicolumn{3}{c}{Regular} & \multicolumn{6}{c}{Irregular}            & \multicolumn{2}{c}{Occluded} & \multicolumn{1}{c}{\multirow{2}{*}{Avg.}}  \\
\cmidrule(lr){3-5} \cmidrule(lr){6-11} \cmidrule(lr){12-13}
\multicolumn{1}{l}{}                                & \multicolumn{1}{c}{}                        & IIIT    & SVT     & IC13    & IC15 & SVTP & CUTE & COCO & CTW  & TT   & HOST & WOST                      &                       \\ \hline
\multirow{4}{*}{1\% (27.8K)}                                & \multicolumn{1}{l|}{Scratch-ViT-Small}  & 12.6  & 3.9  & 10.3 & 7.56 &  3.41 &  6.9 &  2.2 &  4.6 &  4.5 &  5.4 &  \multicolumn{1}{c|}{6.0}    & 5.2                   \\
                                                    & \multicolumn{1}{l|}{Gen-ViT-Small}      & 87.2  & 84.9 & 89.5 & 76.0 & 75.5 &  72.6 & 52.0 & 63.7 & 64.7 & 30.1 & \multicolumn{1}{c|}{\textbf{54.2}}   & 60.6                  \\
                                                    & \multicolumn{1}{l|}{Dis-ViT-Small}      & 87.5  & 85.9 & 88.9 & 75.9 & 73.3 & 72.9 & 52.8 & 63.9 & 65.2 &  30.3 &  \multicolumn{1}{c|}{49.5}   & 60.6                  \\
                                                    & \multicolumn{1}{l|}{DiG-ViT-Small}      & \textbf{88.4}  & \textbf{86.2} & \textbf{89.9} & \textbf{79.0} & \textbf{76.6} & \textbf{77.8} & \textbf{54.8} & \textbf{67.9} & \textbf{67.2} & \textbf{33.2} & \multicolumn{1}{c|}{53.3}   & \textbf{62.9}                  \\ \hline
\multirow{4}{*}{10\% (278K)}                               & \multicolumn{1}{l|}{Scratch-ViT-Small}  & 78.4  & 73.6 & 81.8 & 66.8 & 64.8 & 56.6 &  43.2 & 48.9 & 54.4 & 30.7 & \multicolumn{1}{c|}{48.4}   & 52.3                  \\
                                                    & \multicolumn{1}{l|}{Gen-ViT-Small}      & 95.0  & 92.3 & 95.1 & 83.7 & 84.7 & 90.6 & 65.1 & 79.3 & 80.6 & 37.8 & \multicolumn{1}{c|}{63.9}   & 71.9                  \\
                                                    & \multicolumn{1}{l|}{Dis-ViT-Small}      & 94.6  & 92.3 & 94.6 & 84.5 & 86.2 &  89.9 & 65.7 & 78.2 & 79.8 & 39.0 & \multicolumn{1}{c|}{61.3}   & 71.9                  \\
                                                    & \multicolumn{1}{l|}{DiG-ViT-Small}      & \textbf{95.3}  & \textbf{94.4} & \textbf{95.9} & \textbf{85.3} & \textbf{87.9} & \textbf{91.7} & \textbf{67.1} & \textbf{80.5} & \textbf{81.1} &  \textbf{42.1} & \multicolumn{1}{c|}{\textbf{64.0}}   & \textbf{73.5}                  \\ \hline
\multirow{4}{*}{100\% (2.78M)}                              & \multicolumn{1}{l|}{Scratch-ViT-Small}  & 95.0  & 92.9 & 94.9 & 85.2 &  86.7 & 88.9 & 66.1 &  78.8 & 81.0 & 44.8 & \multicolumn{1}{c|}{67.9}   & 73.4                  \\
                                                    & \multicolumn{1}{l|}{Gen-ViT-Small}      & 97.2  & \textbf{97.1} & \textbf{97.6} & 88.5 & 91.5 & 95.5 & 74.6 & 86.0 & \textbf{89.2} & 54.4 & \multicolumn{1}{c|}{74.3}   & 80.2                  \\
                                                    & \multicolumn{1}{l|}{Dis-ViT-Small}      & 97.1  & 95.7 & 97.4 & 88.1 & \textbf{92.1} & 94.8 & 74.3 & 85.2 & 88.7 & 55.5 &  \multicolumn{1}{c|}{72.9}   & 79.9                  \\
                                                    & \multicolumn{1}{l|}{DiG-ViT-Small}      & \textbf{97.7}  & 96.1 & 97.3 & \textbf{88.6} & 91.6 & \textbf{96.2} & \textbf{75.0} & \textbf{86.3} & 88.9 & \textbf{56.0} &  \multicolumn{1}{c|}{\textbf{75.7}}   & \textbf{80.7}                  \\ \hline
\end{tabular}
}
\end{table*}

\subsection{Implementation Details}
\noindent \textbf{Self-Supervised Pre-Training} We use a vanilla ViT as the feature encoder. To be comparable with various state-of-the-art text recognizers with different model sizes, we adopt 3 ViT variants, namely ViT-Tiny, ViT-Small, and ViT-Base. The only difference between these variants is the embedding size and 192, 384, and 512 are used respectively. To reduce experimental overhead, ViT-Small is adopted as the default backbone in the ablation study. Both unlabeled real data (CC-OCR) and synthetic text data (SynthText and Synth90k) without labels are used for pre-training. We employ an AdamW optimizer~\cite{adamw} with a \textit{cosine} learning rate scheduler, and train the models for 3 epochs. The training hyperparameters are: the batch size as 1,024, base learning rate as 1.5e-4, weight decay as 0.05, $\beta_1=\text{0.9}$, $\beta_2=\text{0.95}$, warm-up for 5,000 steps. All experiments are conducted with 8/16 NVIDIA V100 (32GB RAM) GPUs.

\noindent \textbf{Text Recognition Fine-Tuning} For CTC and Attention decoder, we inherit the configurations from SeqCLR~\cite{SeqCLR} and PerSec~\cite{liu2022perceiving}. As for the transformer decoder, we follow the settings of SATRN~\cite{satrn}, where a 6-layer transformer block with embedding dimension 512 is used. Except for Table 3, where all kinds of decoders are exhaustively discussed, the recognition decoder is based on transformer layers by default. We still employ an AdamW optimizer and a \textit{cosine} learning rate scheduler. The fine-tuning hyperparameters are: the batch size as 2,048, a base learning rate of 1e-4, a weight decay of 0.05, $\beta_1=\text{0.9}$, $\beta_2=\text{0.999}$, warm-up for one epoch. Following ABINet~\cite{abinet}, we train the network for 10 epochs and employ the same image augmentations, including brightness adjustment, noise disturbance, and perspective distortion.

\noindent \textbf{Text Segmentation Fine-Tuning}
The configuration of fine-tuning is similar to the pre-training period except that all patches of the input are kept. Inspired by~\cite{Touvron2022ThreeTE}, for the encoder, we only fine-tune the parameters of self-attention parts to avoid overfitting. The batch size is 256 and the model is trained for 300 epochs. Note that TextSeg~\cite{textseg} is proposed for text segmentation on the whole image, while our task is performed on the text patch. Therefore, we firstly obtain text patches based on the bounding box annotations, and only one text instance is kept in each patch. In both training and evaluation periods, the text patches are resized to $32 \times 128$.

\noindent \textbf{Text Image Super-Resolution Fine-Tuning}
In the previous methods, the resolution of input images is usually smaller than the outputs. 
To be consistent with the pipeline of DiG pre-training, where the inputs and outputs have the same size, we first rescale the input images to $32 \times 128$ with BICUBIC sampling, and then feed them to our super-resolution model.
The configuration of fine-tuning is similar to the text segmentation task. The batch size is 512 and training for 200 epochs. 
Following previous works~\cite{Dong2016ImageSU,Ledig2017PhotoRealisticSI,textzoom,Chen2021SceneTT}, Peak Signal-to-Noise Ratio (PSNR) and structural Similarity Index Measure (SSIM)~\citep{ZhouWang2004ImageQA} are used to evaluate the quality of super-resolution images, and the implementation is based on the official code~\footnote{\url{https://github.com/JasonBoy1/TextZoom}}.

\begin{figure*}[ht]
\centering
\includegraphics[width=0.95\linewidth]{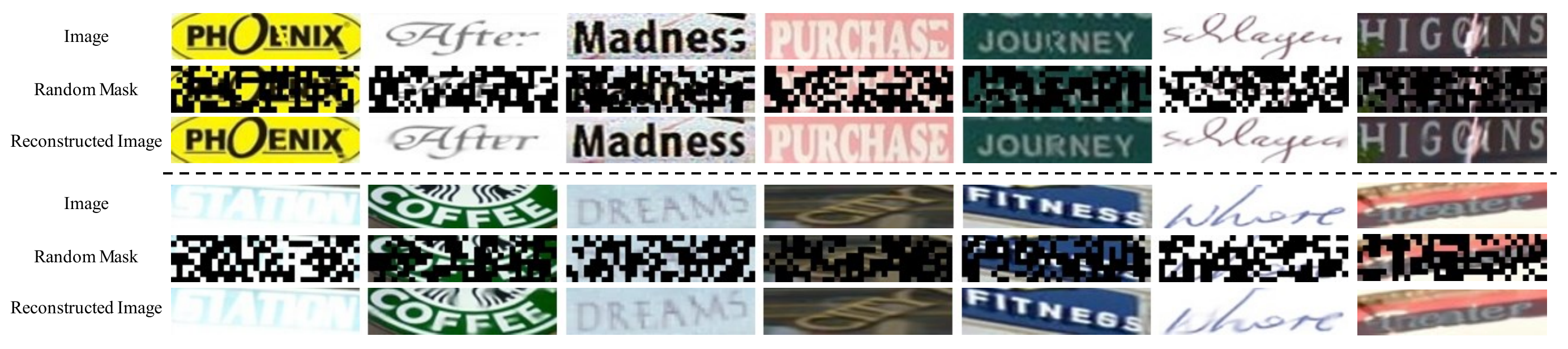}
\caption{Reconstruction results of the masked image modeling.}
\vspace{-4mm}
\label{fig:mim_vis}
\end{figure*}

\subsection{Self-Supervised Learning}

\noindent \textbf{Reconstruction of the Masked Image Modeling} Some reconstruction results of the masked image modeling are visualized in Fig.~\ref{fig:mim_vis}. From the visualization, we can perceive that the reconstruction can produce convincing reconstructions of the text even if the original images are occluded or blurred. This indicates that the model learns outstanding generative feature representations.

\begin{figure}[ht]
\centering
\includegraphics[width=0.98\linewidth]{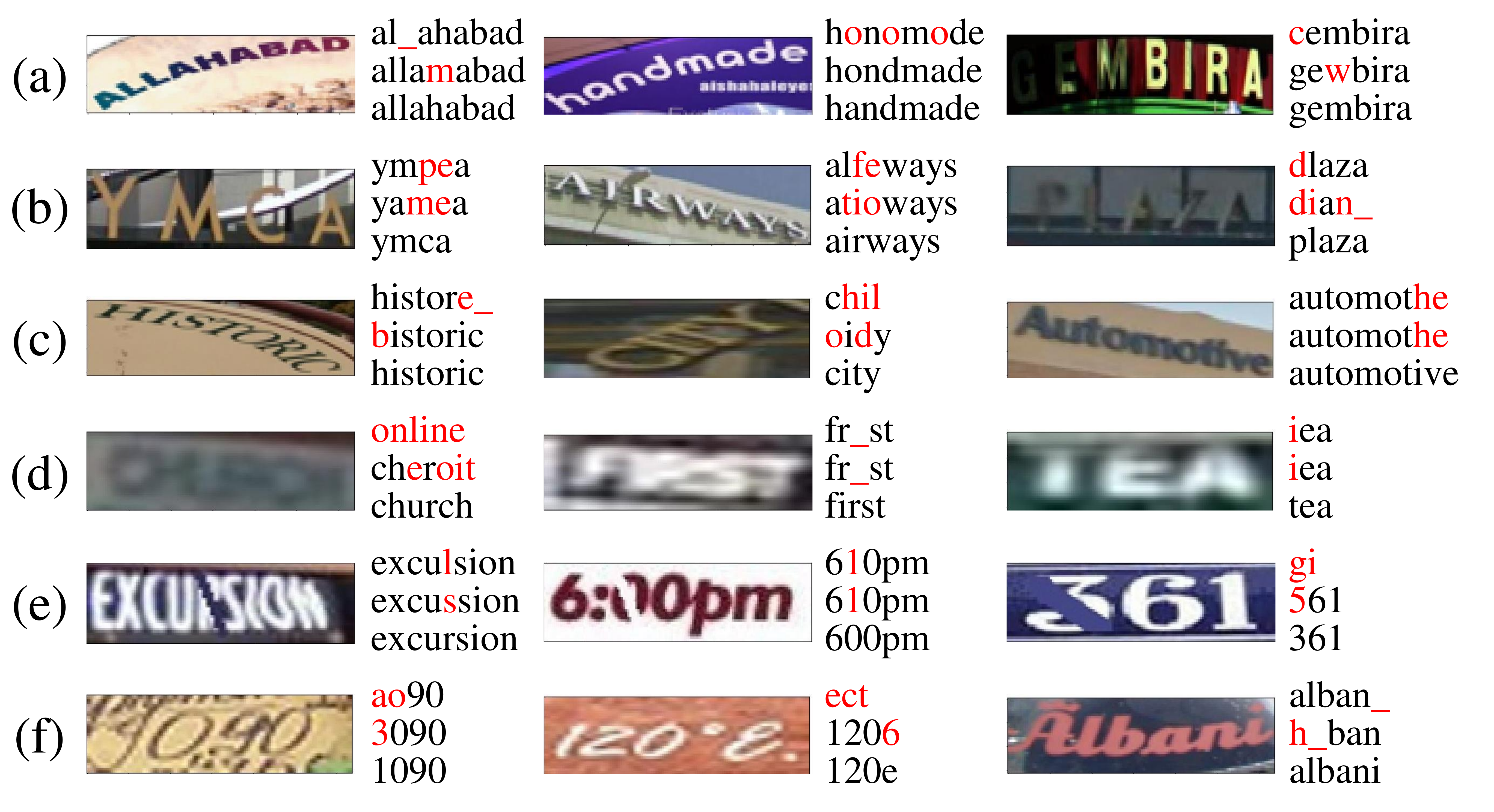}
\caption{Visualization of text recognition results. The three strings near each image represent the prediction of Gen-ViT-Small, Dis-ViT-Small and DiG-ViT-Small respectively, which are fine-tuned with fixed encoder. (a) Curved text. (b) Text disturbed by background. (c) Text with perspective distortion. (d) Blurred text. (e) Text with occlusion. (f) Word art.}
\label{fig:vis_rec}
\end{figure}

\noindent \textbf{Feature Representation Evaluation} We evaluate the quality of the feature representation with different designs by fixing the weights of the encoder and training the text recognition decoder with annotated real data.
As the qualitative comparisons shown in Fig.~\ref{fig:vis_rec}, ``DiG-ViT-Small" performs robust results on various occasions, including curved text, text disturbed by background, text with perspective distortion, blurred text, text with occlusions, and word art, while ``Gen-ViT-Small" and ``Dis-ViT-Small" may make error predictions for these complicated cases.
The quantitative results on standard scene text benchmarks are listed in Tab.~\ref{tab:probe-eval}. Profiting from the integration of discriminative learning and generative learning, ``DiG-ViT-Small" achieves the best results on all benchmarks, surpassing the ``Gen-ViT-Small" and ``Dis-ViT-Small" by 11.8\% and 4.1\% respectively. For the comparison of the discriminative model and the generative model, ``Dis-ViT-Small" obtains higher accuracy on most of the benchmarks while ``Gen-ViT-Small" performs better on the two occluded datasets, due to the similar appearance of occluded images and masked images.

\noindent \textbf{Fine-Tuning Evaluation} We further evaluate our self-supervised model designs by applying them to text recognition models. The text recognition results of different designs and settings are shown in Tab.~\ref{tab:ablation-cl-mim}. ``Gen-ViT-Small", ``Dis-ViT-Small", and ``DiG-ViT-Small" exceed the ``Scratch-ViT-Small" by 6.8\%, 6.5\%, and 7.3\% when training with 100\% of the annotated real data. The improvements go larger with the reduction of the annotated training data. For example, ``DiG-ViT-Small" outperforms ``Baseline-ViT-Small" by 57.7\% and 21.2\% with 1\% and 10\% of the annotated training data. The experimental results show that both contrastive learning and masked image modeling can improve text recognition accuracy and the integration of them achieves the best results, which demonstrates the effectiveness of our proposed DiG.

\begin{table*}[ht]
\caption{Comparisons with existing self-supervised text recognition methods. ``UTI-100M" means using an extra 100M images for pre-training. All the models are trained/fine-tuned with Synthetic Text Data (STD).}
\label{tab:ss-tr}
\resizebox{0.99\textwidth}{!}{
\begin{tabular}{lcccccccccccccc}
\hline
\multicolumn{1}{c}{\multirow{2}{*}{Method}} & \multirow{2}{*}{Decoder}     & \multicolumn{3}{c}{Regular} & \multicolumn{6}{c}{Irregular}            & \multicolumn{2}{c}{Occluded} & \multicolumn{2}{c}{Handwritten} \\
\cmidrule(lr){3-5} \cmidrule(lr){6-11} \cmidrule(lr){12-13} \cmidrule(lr){14-15}
\multicolumn{1}{c}{}                        &                              & IIIT    & SVT     & IC13    & IC15 & SVTP & CUTE & COCO & CTW  & TT    & HOST          & WOST         & IAM            & CVL            \\ \hline
SeqCLR~\cite{SeqCLR}                        & \multirow{5}{*}{CTC}         & 80.9    & -       & 86.3    & -    & -    & -    & -    & -    & -     & -             & -            & 76.7           & 76.9           \\
PerSec-ViT + UTI-100M~\cite{liu2022perceiving}  &                          & 85.4    & 86.1    & 92.8    & 70.3 & 73.9 & 69.2 & -    & -    & -     & -             & -            & 79.9           & 80.5           \\
DiG-ViT-Tiny                                &                              & 93.3    & 89.7    & 92.5    & 79.1 & 78.8 & 83.0 & 58.7 & 69.7 & 72.1  & 32.3          & 53.3         & 79.5           & 82.7           \\
DiG-ViT-Small                               &                              & 95.5    & 91.8    & 95      & 84.1 & 83.9 & 86.5 & 64.3 & 76.0 & 76.87 & 48.6          & 67.7         & 82.7           & 86.4           \\
DiG-ViT-Base                                &                              & \textbf{95.9}    & \textbf{92.6}    & \textbf{95.3}    & \textbf{84.2} & \textbf{85.0} & \textbf{89.2} & \textbf{66.0} & \textbf{77.3} & \textbf{78.7}  & \textbf{58.0}          & \textbf{73.1}         & \textbf{83.2}           & \textbf{87.4}           \\ \hline
SeqCLR~\cite{SeqCLR}                        & \multirow{5}{*}{Attention}   & 82.9    & -       & 87.9    & -    & -    & -    & -    & -    & -     & -             & -            & 79.9           & 77.8           \\
PerSec-ViT + UTI-100M~\cite{liu2022perceiving} &                           & 88.1    & 86.8    & 94.2    & 73.6 & 77.7 & 72.7 & -    & -    & -     & -             & -            & 83.7           & 82.9           \\
DiG-ViT-Tiny                                &                              & 95.1    & 92.4    & 95.8    & 83.2 & 85.4 & 84.7 & 63.8 & 72.3 & 75.9  & 47.7          & 65.1         & 83.8           & 86.6           \\
DiG-ViT-Small                               &                              & 96.4    & \textbf{94.6}    & \textbf{96.6}    & 86.0 & \textbf{89.3} & 88.9 & 68.2 & 76.7 & 80.0  & 65.0          & 77.1         & 84.9           & 89.0           \\
DiG-ViT-Base                                &                              & \textbf{96.8}    & 94.1    & \textbf{96.6}    & \textbf{86.5} & 87.9 & \textbf{92.4} & \textbf{68.7} & \textbf{77.7} & \textbf{81.3}  & \textbf{70.1}          & \textbf{80.2}         & \textbf{85.6}           & \textbf{90.2}           \\ \hline
DiG-ViT-Tiny                                & \multirow{3}{*}{Transformer} & 95.8    & 92.9    & 96.4    & 84.8 & 87.4 & 86.1 & 66.8 & 75.3 & 78.1  & 60.9          & 73.0         & 85.2           & 88.9           \\
DiG-ViT-Small                               &                              & \textbf{96.7}    & 93.4    & \textbf{97.1}    & \textbf{87.1} & 90.1 & 88.5 & 68.8 & 78.8 & 81.1  & 72.1          & 81.1         & 85.7           & 90.5           \\
DiG-ViT-Base                                &                              & \textbf{96.7}    & \textbf{94.6}    & 96.9    & \textbf{87.1} & \textbf{91.0} & \textbf{91.3} & \textbf{69.8} & \textbf{79.3} & \textbf{81.9}  & \textbf{74.9}          & \textbf{82.3}         & \textbf{87.0}           & \textbf{91.3}           \\ \hline
\end{tabular}
}
\end{table*}

\subsection{Text Recognition}
\noindent \textbf{Self-Supervised Text Recognition} We compare the proposed text recognition models with existing self-supervised text recognizers in Tab.~\ref{tab:ss-tr}. All the models in Tab.~\ref{tab:ss-tr} are trained/fine-tuned with the same synthetic text data for fair comparisons. Different decoders are applied for detailed comparisons and the conclusions are similar: our proposed DiG significantly surpasses SeqCLR and PerSec on different benchmarks, including the regular/irregular scene text benchmarks and handwritten text benchmarks. The improvements are especially large on challenging benchmarks, such as IC15, SVTP, and CUTE. For example, ``DiG-ViT-Tiny" outperforms ``PerSec-ViT + UTI-100M" by 8.8\%, 4.9\%, and 13.8\% on IC15, SVTP, and CUTE, with the CTC decoder. The gaps respectively grow to 13.9\%, 11.1\%, and 20.0\% if ViT-base is adopted in our model. Note that ``PerSec-ViT + UTI-100M" uses 100M private unlabeled images for pre-training while DiG only uses 15.77M unlabeled images.
Besides, our proposed DiG can be further improved by using a Transformer decoder, which is the default setting in other experiments.

\begin{table}[ht]
\caption{Comparisons on different annotated data. The values in the table are the average accuracy of all the scene text recognition benchmarks.}
\label{tab:train_data}
\begin{tabular}{cccc}
\hline
Annotated Data & \begin{tabular}[c]{@{}c@{}}Scratch-\\ ViT-Tiny\end{tabular} & \begin{tabular}[c]{@{}c@{}}Scratch-\\ ViT-Small\end{tabular} & \begin{tabular}[c]{@{}c@{}}Scratch-\\ ViT-Base\end{tabular} \\ \hline
STD                                                & \textbf{68.7}                                                        & \textbf{73.6}                                                         & 75.3                                                        \\
ARD                                              & 65.5                                                        & 73.4                                                         & \textbf{76.4}                                                       \\ \hline
Annotated Data & \begin{tabular}[c]{@{}c@{}}DiG-\\ ViT-Tiny\end{tabular}     & \begin{tabular}[c]{@{}c@{}}DiG-\\ ViT-Small\end{tabular}     & \begin{tabular}[c]{@{}c@{}}DiG-\\ ViT-Base\end{tabular}     \\ \hline
STD                                                & 75.3                                                        & 78.8                                                         & 79.7                                                        \\
ARD                                               & \textbf{77.0}                                                       & \textbf{80.7}                                                         & \textbf{82.2}                                                       \\ \hline
\end{tabular}
\end{table}

\noindent \textbf{Synthetic Text Data (STD) \textit{vs.} Annotated Real Data (ARD)} We compare different annotated training data in Tab.~\ref{tab:train_data}. The STD (17M) provides more semantic information due to its large scale while the ARD (2.78M) contains more variety in appearance. Two observations can be perceived as follows:

(1) For the models trained from scratch, small models (``Scratch-ViT-Tiny" and ``Scratch-ViT-Small") achieve higher accuracy with the STD while large models (e.g. ``Scratch-ViT-Base") obtain better results with the ARD. The reason is probably that the small models are limited to appearance representation thus these models benefit more from the rich semantic information in the STD. When the model goes larger, stronger appearance representations help the model obtain higher results with the ARD.

(2) For the models with DiG pre-training, the models fine-tuned with the ARD consistently achieve better results than those fine-tuned with the STD, which is different from the small models trained from scratch. This indicates that the DiG pre-training alleviates the issue of limited appearance representation.

The first observation tells us the reason that previous scene text recognition methods are mostly trained with the STD instead of annotated real data, to a certain extent. With the proposed DiG, using the ARD can achieve consistently higher results, even if the scale of the ARD is smaller than the STD. Thus, DiG can not only sufficiently utilize unlabeled images but also inspire the potential of the annotated real data for text recognition.

\begin{figure}[ht]
\includegraphics[scale=0.35]{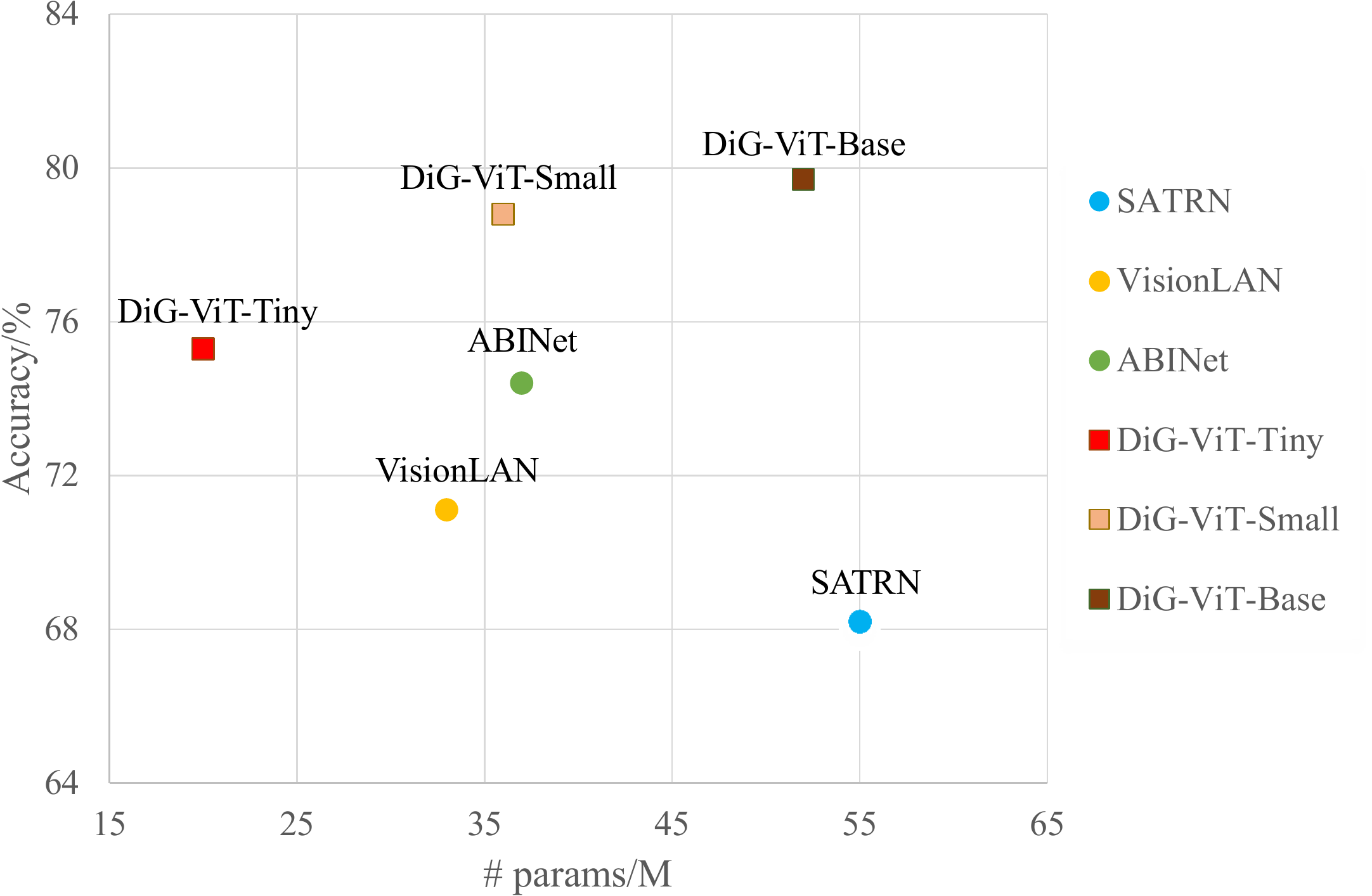}
\caption{Comparisons of number of model parameters and average accuracy on scene text recognition benchmarks.}
\vspace{-4mm}
\label{fig:tradeoff}
\end{figure}

\begin{table*}[ht]
\caption{Scene text recognition results on standard benchmarks. ``*" indicates the results are tested with the officially released models. ``WiKi" indicates using a language model trained with data from WiKiText-103~\cite{wiki}.}
\label{tab:sota-tr}
\resizebox{0.99\textwidth}{!}{
\begin{tabular}{llccccccccccccc}
\hline
\multicolumn{1}{c}{\multirow{2}{*}{Method}} & \multicolumn{1}{c}{\multirow{2}{*}{\begin{tabular}[c]{@{}c@{}}Annotaed\\ Data\end{tabular}}} & \multicolumn{3}{c}{Regular} & \multicolumn{6}{c}{Irregular}           & \multicolumn{2}{c}{Occluded}     & \multirow{2}{*}{Avg.}     & \multirow{2}{*}{Params.} \\
\cmidrule(lr){3-5} \cmidrule(lr){6-11} \cmidrule(lr){12-13}
\multicolumn{1}{c}{}                        & \multicolumn{1}{c}{}                                                                      & IIIT    & SVT     & IC13    & IC15 & SVTP & CUTE & COCO & CTW  & TT   & HOST & WOST                      &                           &                          \\ \hline
\multicolumn{1}{l|}{SE-ASTER~\cite{SE_ASTER}}               & \multicolumn{1}{c|}{STD}                                                  & 93.8    & 89.6    & -       & 80   & 81.4 & 84.4 & -    & -    & -    & -    & \multicolumn{1}{c|}{-}    & \multicolumn{1}{c|}{-}    & -                        \\
\multicolumn{1}{l|}{Textscanner~\cite{TextScanner}}  & \multicolumn{1}{c|}{STD}                                                         & 93.9    & 90.1    & 92.9    & 79.4 & 84.3 & 83.3 & -    & -    & -    & -    & \multicolumn{1}{c|}{-}    & \multicolumn{1}{c|}{-}    & -                        \\
\multicolumn{1}{l|}{DAN~\cite{DAN}}                    & \multicolumn{1}{c|}{STD}                                                       & 94.3    & 89.2    & -       & -    & 80.0 & 84.4 & -    & -    & -    & -    & \multicolumn{1}{c|}{-}    & \multicolumn{1}{c|}{-}    & -                        \\
\multicolumn{1}{l|}{RobustScanner~\cite{RobustScanner}}          & \multicolumn{1}{c|}{STD}                                             & 95.3    & 88.1    & -       & -    & 79.5 & 90.3 & -    & -    & -    & -    & \multicolumn{1}{c|}{-}    & \multicolumn{1}{c|}{-}    & -                        \\
\multicolumn{1}{l|}{AutoSTR~\cite{autostr}}                & \multicolumn{1}{c|}{STD}                                                   & 94.7    & 90.9    & -       & 81.8 & 81.7 & -    & -    & -    & -    & -    & \multicolumn{1}{c|}{-}    & \multicolumn{1}{c|}{-}    & -                        \\
\multicolumn{1}{l|}{SRN~\cite{SRN}}                    & \multicolumn{1}{c|}{STD}                                                       & 94.8    & 91.5    & 95.5    & 82.7 & 85.1 & 87.8 & -    & -    & -    & -    & \multicolumn{1}{c|}{-}    & \multicolumn{1}{c|}{-}    & -                        \\
\multicolumn{1}{l|}{GA-SPIN~\cite{GA-SPIN}}                & \multicolumn{1}{c|}{STD}                                                   & 95.2    & 90.9    & -       & 82.8 & 83.2 & 87.5 & -    & -    & -    & -    & \multicolumn{1}{c|}{-}    & \multicolumn{1}{c|}{-}    & -                        \\
\multicolumn{1}{l|}{Mask TextSpotter~\cite{LiaoLHYWB21}}                & \multicolumn{1}{c|}{STD}                                                   & 95.3    & 91.8    & 95.3       & 78.2 & 83.6 & 88.5 & -    & -    & -    & -    & \multicolumn{1}{c|}{-}    & \multicolumn{1}{c|}{-}    & -                        \\
\multicolumn{1}{l|}{PREN2D~\cite{Pren2D}}                 & \multicolumn{1}{c|}{STD}                                                    & 95.6    & 94      & 96.4    & 83   & 87.6 & 91.7 & -    & -    & -    & -    & \multicolumn{1}{c|}{-}    & \multicolumn{1}{c|}{-}    & -                        \\
\multicolumn{1}{l|}{JVSR~\cite{JVSR}}                   & \multicolumn{1}{c|}{STD}                                                      & 95.2    & 92.2    & -       & -    & 85.7 & 89.7 & -    & -    & -    & -    & \multicolumn{1}{c|}{-}    & \multicolumn{1}{c|}{-}    & -                        \\
\multicolumn{1}{l|}{SATRN~\cite{satrn}}                  & \multicolumn{1}{c|}{STD}                                                     & 92.8    & 91.3    & -       & -    & 86.5 & 87.8 & -    & -    & -    & -    & \multicolumn{1}{c|}{-}    & \multicolumn{1}{c|}{-}    & 55M                      \\
\multicolumn{1}{l|}{SATRN*}                 & \multicolumn{1}{c|}{STD}                                                                  & 92.5    & 93.9    & 96.3    & 81.3 & 86.5 & 85.8 & 50.4 & 70.8 & 74   & 63.4 & \multicolumn{1}{c|}{76.3} & \multicolumn{1}{c|}{68.2} & 55M                      \\
\multicolumn{1}{l|}{VisionLAN~\cite{ost}}              & \multicolumn{1}{c|}{STD}                                                       & 95.8    & 91.7    & 95.7    & 83.7 & 86.0 & 88.5 & -    & -    & -    & -    & \multicolumn{1}{c|}{-}    & \multicolumn{1}{c|}{-}    & 33M                      \\
\multicolumn{1}{l|}{VisionLAN*}             & \multicolumn{1}{c|}{STD}                                                                  & 95.9    & 92.0    & 96.3    & 84.1 & 85.9 & 88.9 & 59.2 & 75.1 & 78.7 & 49.8 & \multicolumn{1}{c|}{70.8} & \multicolumn{1}{c|}{71.1} & 33M                      \\
\multicolumn{1}{l|}{ABINet~\cite{abinet}}                 & \multicolumn{1}{c|}{STD+WiKi}                                               & 96.2    & 93.5    & 97.4    & 86.0 & 89.3 & 89.2 & -    & -    & -    & -    & \multicolumn{1}{c|}{-}    & \multicolumn{1}{c|}{-}    & 37M                      \\
\multicolumn{1}{l|}{ABINet*}                & \multicolumn{1}{c|}{STD+WiKi}                                                             & 96.4    & 94.4    & 97.0    & 85.9 & 89.6 & 88.5 & 63   & 76.8 & 80.7 & 57.9 & \multicolumn{1}{c|}{75.3} & \multicolumn{1}{c|}{74.4} & 37M                      \\ \hline
\multicolumn{1}{l|}{DiG-ViT-Tiny}           & \multicolumn{1}{c|}{STD}                                                                  & 95.8    & 92.9    & 96.4    & 84.8 & 87.4 & 86.1 & 66.8 & 75.3 & 78.1 & 60.9 & \multicolumn{1}{c|}{73.0} & \multicolumn{1}{c|}{75.3} & 20M                      \\
\multicolumn{1}{l|}{DiG-ViT-Small}          & \multicolumn{1}{c|}{STD}                                                                  & 96.7    & 93.4    & 97.1    & 87.1 & 90.1 & 88.5 & 68.8 & 78.8 & 81.1 & 72.1 & \multicolumn{1}{c|}{81.1} & \multicolumn{1}{c|}{78.8} & 36M                      \\
\multicolumn{1}{l|}{DiG-ViT-Base}           & \multicolumn{1}{c|}{STD}                                                                  & 96.7    & 94.6    & 96.9    & 87.1 & 91.0 & 91.3 & 69.7 & 79.3 & 81.9 & \textbf{74.9} & \multicolumn{1}{c|}{\textbf{82.3}} & \multicolumn{1}{c|}{79.7} & 52M                      \\ \hline
\multicolumn{1}{l|}{DiG-ViT-Tiny}          & \multicolumn{1}{c|}{ARD}                                                   & 96.4    & 94.4    & 96.2    & 87.4 & 90.2 & 94.1 & 71.8 & 83.1 & 86.6 & 45.3 & \multicolumn{1}{c|}{68.2} & \multicolumn{1}{c|}{77.0} & 20M                      \\
\multicolumn{1}{l|}{DiG-ViT-Small}         & \multicolumn{1}{c|}{ARD}                                                   & \textbf{97.7}    & 96.1    & 97.3    & 88.6 & 91.6 & 96.2 & 75.0 & 86.3 & 88.9 & 56.0 & \multicolumn{1}{c|}{75.7} & \multicolumn{1}{c|}{80.7} & 36M                      \\
\multicolumn{1}{l|}{DiG-ViT-Base}          & \multicolumn{1}{c|}{ARD}                                                   & 97.6    & \textbf{96.5}    & \textbf{97.6}    & \textbf{88.9} & \textbf{92.9} & \textbf{96.5} & \textbf{75.8} & \textbf{87.0} & \textbf{90.1} & 62.8 & \multicolumn{1}{c|}{79.7} & \multicolumn{1}{c|}{\textbf{82.2}} & 52M                      \\ \hline
\end{tabular}
}
\end{table*}

\noindent \textbf{Scene Text Recognition} We depict the number of model parameters and the average accuracy on scene text recognition benchmarks of state-of-the-art methods in Fig.~\ref{fig:tradeoff}. It clearly shows that our proposed DiG achieves a better trade-off between the model size and the accuracy. For example, our ``DiG-ViT-Tiny" outperforms all previous methods with the minimal model size. More experimental results are shown in Tab.~\ref{tab:sota-tr}. With the assistance of our proposed DiG, our text recognition models significantly surpass previous state-of-the-art text recognition methods on 11 scene text recognition benchmarks. Specifically, even if ABINet~\cite{abinet} adopts an extra language model that is trained with WiKi data, our ``DiG-ViT-Small" outperforms it by 4.4\% on average accuracy of the scene text benchmarks, with almost the same model size (37M \textit{vs.} 36M). Compared to SATRN~\cite{satrn} that also applies a ViT encoder and a Transformer decoder, our ``DiG-ViT-Base" significantly surpasses it by 11.5\% on average, with approximate model size (52M \textit{vs.} 55M).

As discussed above, the annotated real data can achieve better results with the strong feature representation learned from DiG. Therefore, we also list the results fine-tuned with ARD data in Tab.~\ref{tab:sota-tr}, which further refreshes the state-of-the-art results on most of the scene text recognition benchmarks.

\subsection{Text Segmentation}
\begin{figure}[ht]
\centering
\captionsetup[subfigure]{justification=centering}
\begin{subfigure}[b]{0.48\textwidth}
         \centering
         \includegraphics[width=\textwidth]{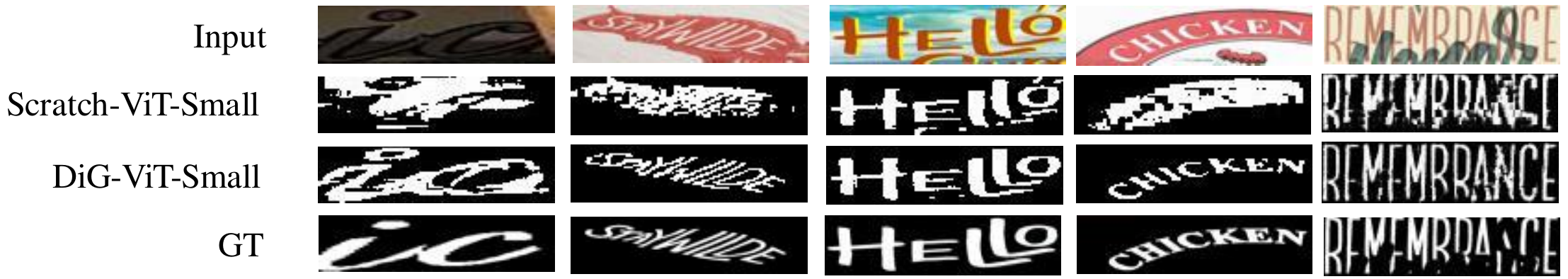}
         \caption{}
         \label{subfig:vis_seg}
\end{subfigure}
\begin{subfigure}[b]{0.48\textwidth}
         \centering
         \includegraphics[width=\textwidth]{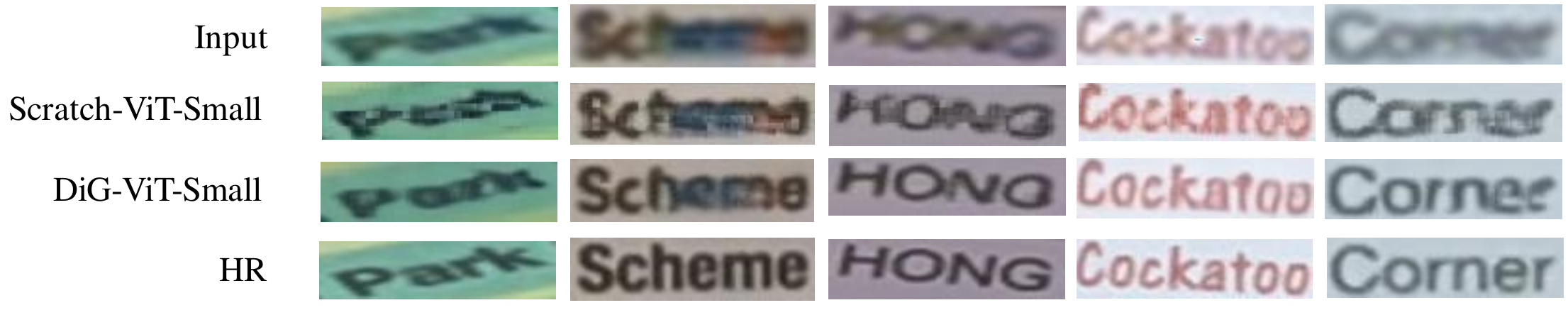}
         \caption{}
         \label{subfig:vis_sr}
\end{subfigure}
\caption{Illustration of (a) the text segmentation results and (b) text image super-resolution results. ``HR" indicates the provided high-resolution images. Best view with zoom in.}
\label{fig:vis_sr_seg}
\end{figure}

We are the first to conduct text segmentation experiments at patch level on the TextSeg dataset. Thus, we compare our text segmentation model with self-supervised pre-training to the baseline that is trained from scratch. As shown in Tab.~\ref{tab:text_seg}, ``DiG-ViT-Small" outperforms ``Scratch-ViT-Small" by 5\% with the IoU metric. Some visualized text segmentation results are depicted in Fig.~\ref{subfig:vis_seg}. We can observe that ``DiG-ViT-Small" performs better segmentation effects than ``Scratch-ViT-Small" on complex occasions where text images suffer from background noise, perspective distortion, or low resolution.

\begin{table}[ht]
\caption{Text segmentation results on the TextSeg dataset.}
\label{tab:text_seg}
\begin{tabular}{ll}
\hline
Method             & IoU    \\ \hline
Scratch-ViT-Small  & 78.1 \\
DiG-ViT-Small      & \textbf{83.1} \\ \hline
\end{tabular}
\end{table}

\begin{table}[ht]
\caption{Text image super-resolution results on the TextZoom dataset. All methods use L2 loss for fair comparisons.}
\label{tab:Text_super}
\resizebox{0.49\textwidth}{!}{
\begin{tabular}{lcccccc}
\hline
\multirow{2}{*}{Method}  & \multicolumn{3}{c}{SSIM} & \multicolumn{3}{c}{PSNR} \\ \cline{2-7} 
                                              & Easy   & Medium & Hard   & Easy   & Medium  & Hard  \\ \hline
SRCNN~\cite{Dong2016ImageSU}                  & 0.8152 & 0.6425 & 0.6833 & 23.13  & 19.57   & 19.56 \\
SRResNet~\cite{Ledig2017PhotoRealisticSI}     & 0.8176 & 0.6324 & 0.7060 & 20.65  & 18.90   & 19.50 \\
TSRN~\cite{textzoom}                   & 0.8562 & 0.6596 & 0.7285 & 22.95  & 19.26   & 19.76 \\ 
TBSRN~\cite{Chen2021SceneTT}                  & 0.8729 & 0.6455 & 0.7452 & 24.13  & 19.08   & 20.09 \\ \hline
Scratch-ViT-Small                             & 0.8143 & 0.6288 & 0.6845 & 22.90  & 19.65   & 20.45 \\
DiG-ViT-Small                                 & 0.8613 & 0.6561 & 0.7215 & 23.98  & 19.85   & 20.57 \\ \hline
\end{tabular}
}
\vspace{-4mm}
\end{table}

\subsection{Text Image Super-Resolution}
Our foundation models can be also applied to text image super-resolution models and obtain significant performance gain. Some results on the TextZoom dataset are illustrated in Fig.~\ref{subfig:vis_sr}. ``DiG-ViT-Small" produces better super-resolution images than ``Scratch-ViT-Small". The quantitative results are shown in Tab.~\ref{tab:Text_super}. Benefited from DiG, ``DiG-ViT-Small" exceeds ``Scratch-ViT-Small" on both SSIM evaluation and PSNR evaluation. Besides, our proposed model achieves competitive even better results than existing state-of-the-art methods, without any specific designs. 

\subsection{Limitations}
Masked image modeling is designed for ViT-based models and has not been verified on CNNs. Thus, our proposed DiG may be not suitable for CNN backbones. However, this is not a big issue since most existing text recognizers are sequence-to-sequence models, which exactly match the structure of ViTs or Transformers.

\section{Conclusion}
In this paper, we have investigated self-supervised learning for text recognition. The proposed DiG is an integration of contrastive learning and masked image modeling, which gains the superiority of both discriminative representation and generative representation. We validate DiG on text recognition and other text-related tasks such as text image super-resolution and text segmentation. With the foundation model pre-trained by DiG, the performance of the task-specific models can be significantly improved. In the future, we will explore self-supervised learning for the combination of text recognition and natural language processing.

\bibliographystyle{ACM-Reference-Format}
\bibliography{reference}

\end{document}